\documentclass[lettersize,journal]{IEEEtran}
\usepackage{amsmath,amsfonts}
\usepackage{algorithmic}
\usepackage{algorithm}
\usepackage{array}
\usepackage[caption=false,font=normalsize,labelfont=sf,textfont=sf]{subfig}
\usepackage{textcomp}
\usepackage{stfloats}
\usepackage{url}
\usepackage{verbatim}
\usepackage{graphicx}
\usepackage{cite}
\usepackage{booktabs}
\usepackage{color}
\usepackage[colorlinks,linkcolor=blue]{hyperref}

\begin{document}

\title{Magic-Boost: Boost 3D Generation with Multi-View Conditioned Diffusion}

\author{Fan Yang, Jianfeng Zhang, Yichun Shi, Bowen Chen, Chenxu Zhang, Huichao Zhang, Xiaofeng Yang, Xiu Li, Jiashi Feng, Guosheng Lin
\thanks{Corresponding author: Guosheng Lin}
\thanks{Fan Yang, Xiaofeng Yang and Guosheng Lin are with the
College of Computing and Data Science, Nanyang Technological University (NTU),  639798 Singapore (email: fan007@e.ntu.edu.sg,yang.xiaofeng@ntu.edu.sg, gslin@ntu.edu.sg)}
\thanks{Jianfeng Zhang and Jiashi Feng are with the ByteDance inc, 078881 Singapore (email: jianfengzhang@bytedance.com, jshfeng@bytedance.com)
}
\thanks{Yichun Shi and Chenxu Zhang are with ByteDance inc, 95110 San Jose, USA
(email: 
yichun.shi@bytedance.com,chenxuzhang@bytedance.com)}
\thanks{Bowen Chen, Huichao Zhang and Xiu Li are with ByteDance inc, 100098 Beijing, China (email: 
chenbowen.cbw@bytedance.com, zhanghuichao.hc@bytedance.com, lixiu.cv@bytedance.com)}
}

\markboth{Journal of \LaTeX\ Class Files,~Vol.~14, No.~8, August~2021}%
{Shell \MakeLowercase{\textit{et al.}}: A Sample Article Using IEEEtran.cls for IEEE Journals}


\maketitle

\begin{abstract}
Benefiting from the rapid development of 2D diffusion models, 3D content generation has witnessed significant progress. One promising solution is to finetune the pre-trained 2D diffusion models to produce multi-view images and then reconstruct them into 3D assets via feed-forward sparse-view reconstruction models. However, limited by the 3D inconsistency in the generated multi-view images and the low reconstruction resolution of the feed-forward reconstruction models, the generated 3d assets are still limited to incorrect geometries and blurry textures. To address this problem, we present a multi-view based refine method, named Magic-Boost, to further refine the generation results. In detail, we first propose a novel multi-view conditioned diffusion model which extracts 3d prior from the synthesized multi-view images to synthesize high-fidelity novel view images and then introduce a novel iterative-update strategy to adopt it to provide precise guidance to refine the coarse generated results through a fast optimization process. Conditioned on the strong 3d priors extracted from the synthesized multi-view images, Magic-Boost is capable of providing precise optimization guidance that well aligns with the coarse generated 3D assets, enriching the local detail in both geometry and texture within a short time ($\sim15$min). Extensive experiments show Magic-Boost greatly enhances the coarse generated inputs, generates high-quality 3D assets with rich geometric and textural details.
\end{abstract}


\begin{figure*}
\maketitle
\centering
\includegraphics[width=0.9\textwidth]{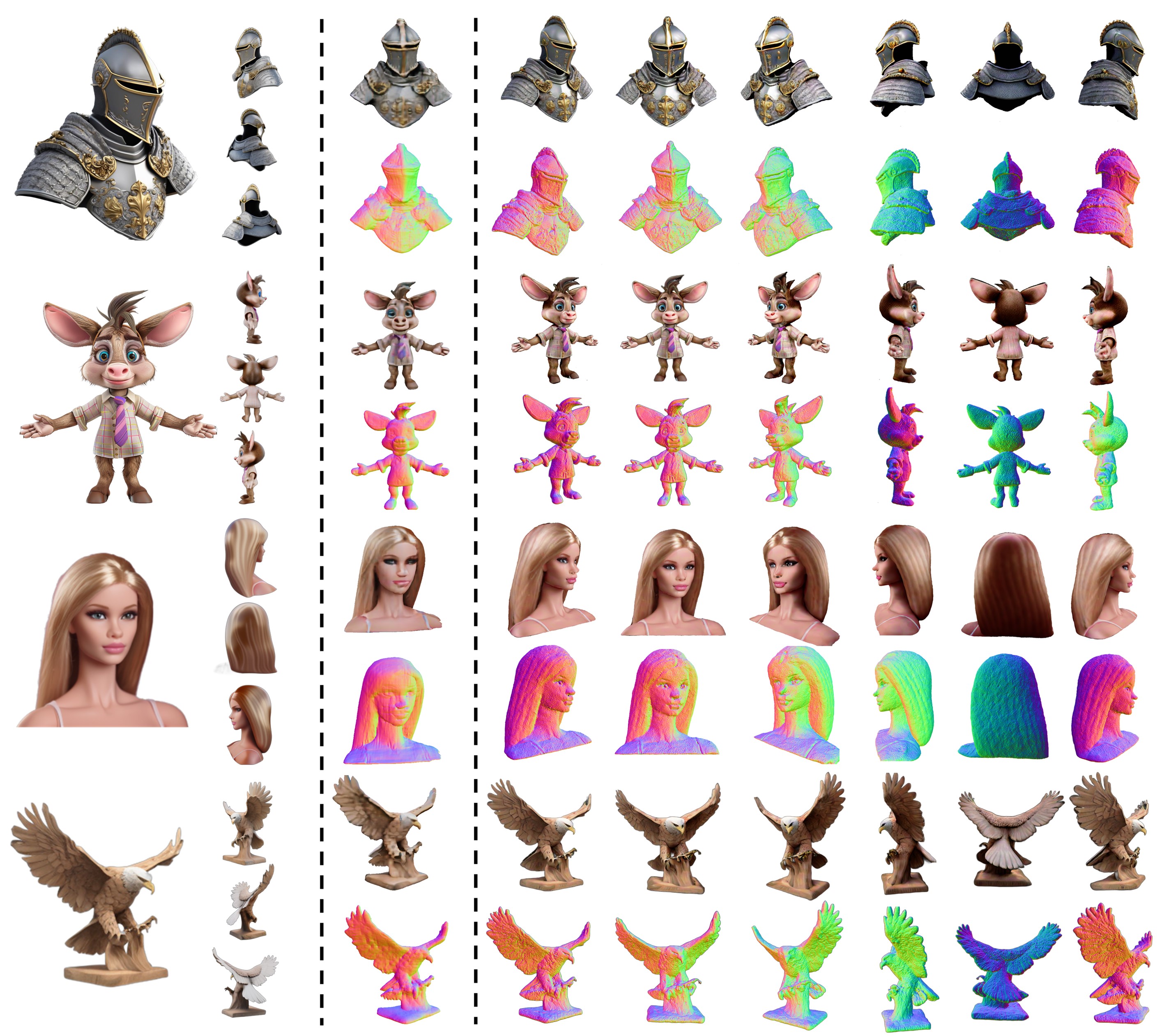}
\caption{Provided with an input image and its  coarse 3D generation, MagicBoost  effectively boosts it to a high-quality 3D asset within 15 minutes. From left to right, we show the input image, pesudo multi-view images and coarse 
3D results from Instant3D~\cite{li2023instant3d}, together with the significantly improved results produced by our method.
}
\label{fig:teaser}
\end{figure*}

\section{Introduction}
\IEEEPARstart{T}{he} recent surge in the development of 2D diffusion models has opened a new door for 3D content generation. 
One of the promising methods is to first synthesize multi-view consistent images with fine-tuned 2D diffusion model and 
then reconstruct them into 3D assets through fast-NeRFs~\cite{liu2023syncdreamer,long2023wonder3D,shi2023zero123++} or large reconstruction models~\cite{li2023instant3d,melas20243D}. 
Despite the efficiency of these methods
, the generated results still suffer from artifacts like coarse textures or incorrect geometries, which is primarily caused by the local inconsistencies existing in the synthesized multi-view images and limited reconstruction resolution from the feed-forward reconstruction models. 

Recently, efforts~\cite{liu2023unidream,ding2023text,qian2023magic123} have been made to adopt SDS (Score Distillation Sampling~\cite{poole2022Dreamfusion}) to further refine the coarse generated results through an optimization process. However, SDS optimization process is hard to control and  leads to instable refinement results with problems like identity shifts, blurred textures, and geometric collapse~\cite{poole2022Dreamfusion,wang2024prolificdreamer,liu2023unidream}. We argue that the main reason for this instability is the lack of 3D understanding and explicit identity control ability of 2D diffusion models on the optimization process. Specifically, the inherent ambiguity of textual descriptions poses a challenge for text-conditioned diffusion models, such as StableDiffusion~\cite{rombach2022high} and DeepFoldy-IF~\cite{saharia2022photorealistic}, to maintain a consistent identity throughout the optimization process. As a result, the refined results may diverge significantly from the initial coarse 3D assets, which potentially defies users' expectations. On the other hand, single-view image conditioned diffusion models like Zero-1-to-3~\cite{liu2023zero} empower 2D diffusion models with the view-condition capacity. However, the information provided by single-view images is very limited, leading to implausible shapes and blurry textures of the refined results.

To address this problem, we introduce a multi-view based refine method, named Magic-Boost, to
further refine the generation results. The main motivation of us is that the synthesized multi-view images, despite not strictly 3D consistent, are still capable of providing strong 3D priors which align well with the target generation results and could be extracted by the multi-view conditioned diffusion model to guide the refinement process. Inspired by this, we first introduce a novel multi-view conditioned diffusion model that adopts synthesized pseudo multi-view images as inputs, implicitly distill 3D information across different views 
to synthesize high-fidelity novel view images and provide precise SDS (Score Distillation Sampling~\cite{poole2022Dreamfusion}) guidance to refine the local details of the coarse generated 3D assets. 
Specifically, we build the model upon the Stable Diffusion architecture~\cite{rombach2022high}, further equip it with a novel time-fixed local feature  extractor to  efficiently capture the low-level local details and enable the interactions across different views with cross-frame 3D attention. During training stage, we meticulously develop a series of data augmentation strategies to imitate the 3D inconsistency to empower the model to extract strong 3D priors from inconsistent multi-view inputs at test time. 
During the optimization process, we present a novel Anchor Iterative Update loss to address the over-saturation problem in the Score Distillation Sampling process~\cite{poole2022Dreamfusion}, culminating in the generation of high-quality content with realistic textures.  

\begin{figure*}[tb]
  \centering
    \includegraphics[width=1.01\linewidth]{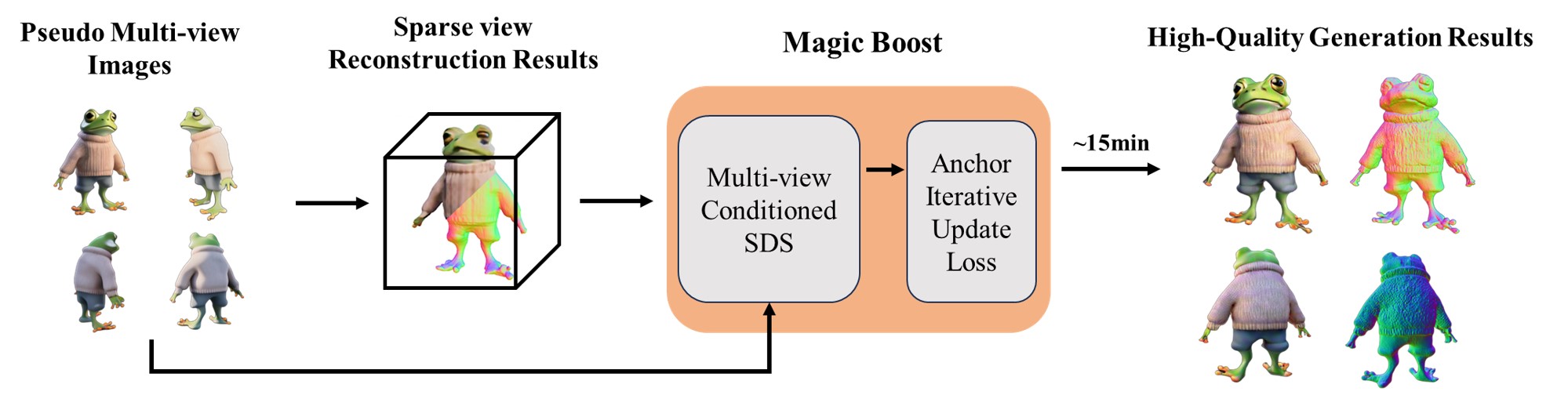}
  \caption{The overall pipeline. The proposed Magic-Boost could be a plug-in module plugged into any 3D generation methods capable of providing
pseudo multi-view priors, such as Instant3D~\cite{li2023instant3d}, InstantMesh~\cite{xu2024instantmesh}, LGM~\cite{tang2024lgm} and etc. Benefit from the strong 3d priors provided by the pesudo multi-view images, Magic-boost provides precise SDS guidance, significantly enhancing the coarse 3D outputs within a brief interval
($\sim15$min).}
  \label{fig:G_pipeline}
\end{figure*}

Figure~\ref{fig:G_pipeline} depicts the overall pipeline of the proposed Magic-Boost refinement method. Magic-Boost could be seen as a plug-in module which could be pluged into any 3D generation methods capable of providing
pseudo multi-view priors, such as Instant3D~\cite{li2023instant3d}, InstantMesh~\cite{xu2024instantmesh}, LGM~\cite{tang2024lgm} and etc. As shown in Figure \ref{fig:teaser}, benefit from the strong 3d priors provided by the pesudo multi-view images, Magic-boost provides precise SDS guidance, significantly enhancing the coarse 3D outputs within a brief interval
($\sim15$min).  Comprehensive evaluations demonstrate Magic-Boost substantially  enhances the quality of coarse inputs, efficiently yielding 3D assets of better quality with intricate geometries and authentic textures.

\section{Related Works}
\label{sec:related}
 \noindent\textbf{3D generation Models.}
Traditional GAN-based methods explore generation of 3D models with different 3D representations including voxels~\cite{henzler2019escaping,sohl2015deep}, point clouds~\cite{yang2019pointflow, achlioptas2018learning,chen2024learning,chen2023sgsr} and meshes~\cite{gao2019sdm,wei2023taps3d,ding2023towards,liu2023transformer}. 
Recently, the advance of diffusion models in 2D generative tasks~\cite{rombach2022high,saharia2022photorealistic,koksal2023controllable} has prompted explorations of their application into 3D domains. Efforts~\cite{jun2023shap} have been made to directly train diffusion models on 3D datasets.  
However, even though the largest 3D dataset~\cite{deitke2023objaverse} in recent years are still
much smaller than the datasets used for 2D image generation
training~\cite{schuhmann2022laion}. Recently,  a  novel paradigm has emerged for 3D generation that circumvents the need for large-scale 3D datasets by leveraging pretrained 2D generative models. Leveraging the semantic understanding and high-quality generation capabilities of pretrained 2D diffusion model, Dreamfusion~\cite{poole2022Dreamfusion}, for the first time, propose to optimize 3D representations directly with 2D diffusion model utilizing the Score distilling Sampling loss. Follwoing works~\cite{qian2023magic123,lin2023magic3d,yang2023learn,chen2024sculpt3d} continue to enhance various aspects such as generation fidelity and training stability by proposing different variations of SDS such as VSD~\cite{wang2024prolificdreamer}. Although being able to generate high-quality 3D contents, these methods suffer from extremely long time for optimization, which greatly hinders its application in the real-world scenario. Another line of works try to finetune the pretrained 2D diffusion network to unlock its ability of generating multi-view images simultaneously, which are subsequently lifted to 3D models with fast-NeRF~\cite{muller2022instant} or large reconstruction models~\cite{liu2023syncdreamer,long2023wonder3D,li2023instant3d,melas20243D}. 
Although these methods generates reasonable results, they are still limited by the local inconsistency and limited generation resolution, producing coarse results without detailed textures and complicated geometries.

\noindent\textbf{Novel View Synthesis.} 
The success of diffusion model on the 2D task has opened a new door for the task of zero-shot novel view synthesis. Finetuned on Stable Diffusion~\cite{rombach2022high}, Zero-1-to-3~\cite{liu2023zero} achieves viewpoint-conditioned image synthesis ability on different inputs. Subsequent~\cite{ye2023consistent,shi2023zero123++,liu2023syncdreamer} works 
further improve the generative quality of Zero-1-to-3 by refining the network architecture and improving the training methodologies. Yet, constrained by the limited information provided by single-view input, these models encounter challenges in producing accurate novel views with robust 3D consistency. In contrast, EscherNet~\cite{kong2024eschernet} propose to 
improve the 3D understanding of the network
by fusing information from arbitrary number of reference and target views, 
leading to better view synthesis and reconstruction results. Compared to EscherNet, our model focus on a more challenging generation task: employing pseudo multi-view images as 3d priors to enhance the accuracy of the 3D generation. 

\section{Methods}
\label{sec:method}

We propose Magic-Boost, a multi-view conditioned diffusion model that explores the strong 3d prior provided by the synthesized multi-view images to refine the coarse generated results.  The proposed model is built upon the Stable Diffusion architecture, where we extract dense local features via a denoising U-Net operating at a fixed timestep and adopt the self-attention mechanism to achieve information exchange across different views, as detailed in Sec.~\ref{sec:MCD}. 
To  empower the
model to extract strong 3D priors from inconsistent multi-
view inputs at test time, we introduce several data augmentation strategies, as elaborated in Sec.~\ref{sec:RT}. In the refinement phase, we introduce an Anchor Iterative Update loss to alleviate the over-saturation problem of SDS~\cite{poole2022Dreamfusion}, as presented in Sec.~\ref{sec:Refine}.

\begin{figure*}[tb]
  \centering
    \includegraphics[width=\linewidth]{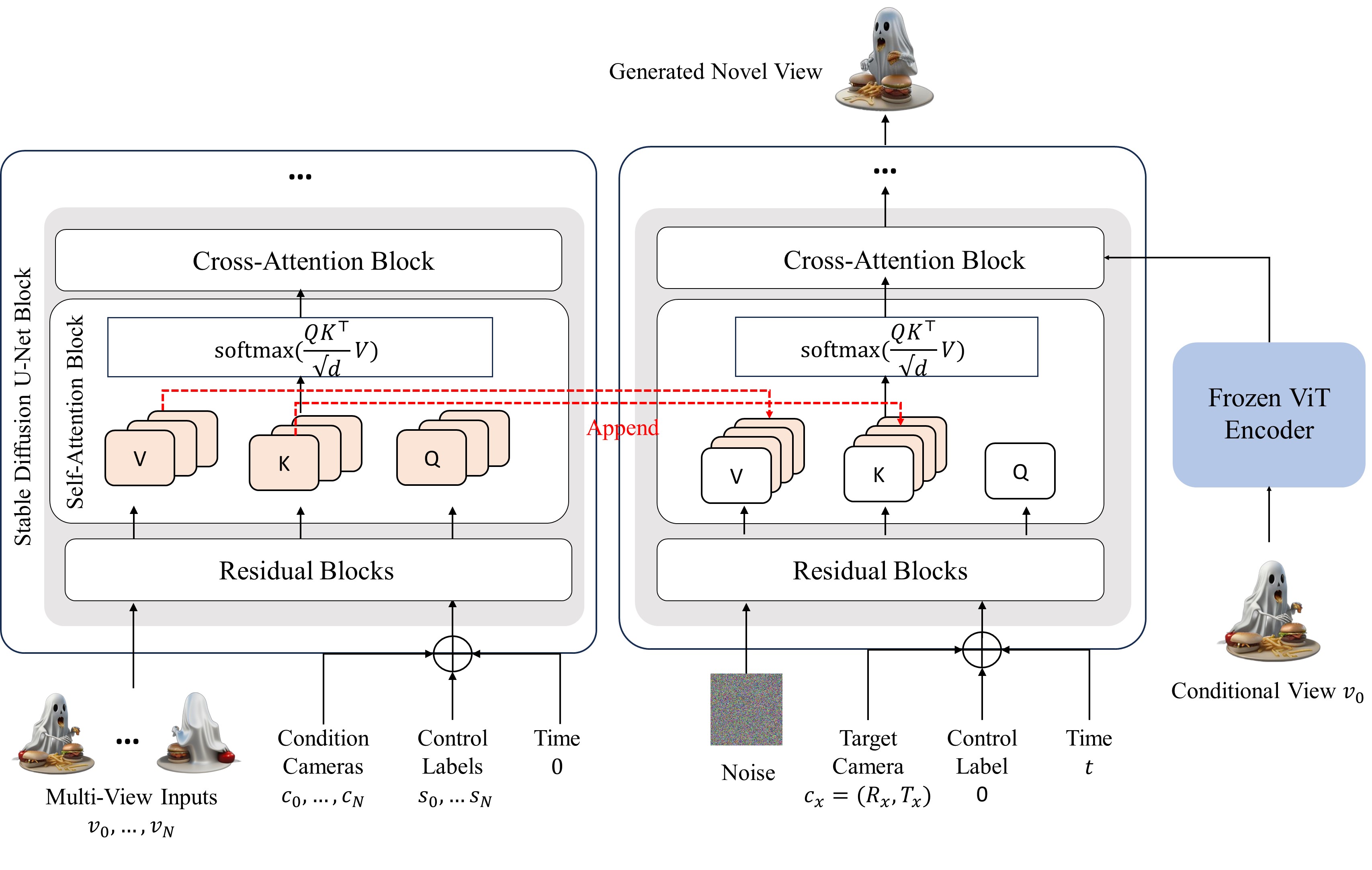}
  \caption{Architecture of our multi-view conditioned diffusion model. At the core of our model lies the extraction of dense local features facilitated by a denoising U-Net operating at a fixed timestep. Concurrently, we harness a frozen CLIP ViT encoder to distill high-level signals. The original 2D self-attention layer is extended into 3D by concatenating keys and values across various views. To further control the condition strength of different views, we involve a control label which allows users to manually control the condition strength of different conditional views.
 }
      \label{fig:network}
\end{figure*}

\subsection{Multi-view Conditioned Diffusion}
\label{sec:MCD}


%
\textbf{Formulation.} 
Given $n$ views $v_0, v_1, ..., v_n \in \mathbb{R}^{H\times W\times 3}$, each capturing an object from a distinct perspective with relative angles $\Delta \gamma_0 = 0, \Delta \gamma_1, ..., \Delta \gamma_n$ as input, the objective of our multi-view conditioned diffusion model is to synthesize a novel view $x$ at a relative angle $\Delta \gamma_{x}$. To this end, we compute the camera  rotation $R_i \in \mathbb{R}^{3\times3}$ and translation $T_i \in \mathbb{R}^3$ corresponding to each relative angle $\Delta \gamma_i, i\in (0,1,...,n,x)$ and subsequently train the model $\mathcal{M}$ to generate the novel view $x$. The relative camera pose is denoted as $c_i = (R_i, T_i), i\in (0,1,...,n,x)$, and the formulation of our model is as follows: 
\[
    x_{c_x} = \mathcal{M}(v_0,...,v_n,c_0,...,c_n,c_x)
\]

In our experiments, we follow the setting of Instant3D~\cite{li2023instant3d} and employing a four-view arrangement, where the four condition views are orthogonal to each other with relative angles set at $\Delta \gamma_0 = 0, \Delta \gamma_1 = 90, \Delta \gamma_2 = 180$, and $ \Delta \gamma_3 = 270$.


\noindent\textbf{Image Feature Extractor.} To synthesize consistent novel views, it's crucial to capture both the global features with high-level semantics, and the local features with dense local details.

\noindent\textit{Global Feature Extractor.} In line with previous works~\cite{liu2023zero,ye2023ip,mou2023t2i,wang2023imagedream}, we utilize a frozen CLIP pre-trained Vision Transformer (ViT)~\cite{radford2021learning} to encode high-level signals, which provide global control on the generated images. However, as CLIP encodes images into a highly compressed feature space, we found that encoding the global feature of only the first input view (where $\gamma_0 = 0$) is sufficient to generate satisfactory results, while encoding multi-view global features does not lead to improvement on the performance. 
Consequently, we only encode the global feature of the first input view in our experiments.

\noindent\textit{Local Feature Extractor.} Multi-view images provide dense local details, pivotal for novel view synthesis. However, accurately and efficiently encoding local dense features from multi-view inputs is non-trivial. Zero-1-to-3~\cite{liu2023zero} attempts to encode dense local signals by appending the reference image to the input of the denoising U-Net within Stable Diffusion. This approach, however, enforces an misaligned pixel-wise spatial correspondence between the input and target images. Zero123++~\cite{shi2023zero123++} extracts local feature by processing an additional reference image through the denoising U-Net model. It synchronizes the Gaussian noise level of the input images with that of the denoising input, enabling the U-Net to focus on pertinent features at varying noise levels. While this method is adept at extracting dense local features, it leads to much more computational costs, by extracting different feature maps at each denoising step. To address this issue, we propose a novel technique that extracts dense local features using the denoising U-Net at a fixed timestep. Specifically, we employ the denoising U-Net to extract low-level features from clean images without introducing any noise, and we consistently set the timestep to zero. 
This approach not only captures features  with dense local details but also significantly accelerates the generation process by extracting local features just once throughout the entire SDS optimization or diffusion procedure.

\noindent\textbf{Multi-view Conditioned Generation.} The overall architecture of our model is shown in Figure~\ref{fig:network}. Our multi-view conditioned diffusion model is built on Stable Diffusion backbone~\cite{rombach2022high}. We incorporate global features via the cross-attention module, following methods such as IP-Adapter~\cite{ye2023ip} and ImageDream~\cite{wang2023imagedream}. The denoising U-Net, operating at a fixed timestep, is utilized to distill dense local features from multi-view inputs. These features are subsequently integrated into the self-attention module to encode 3D correspondence. Similar with~\cite{shi2023mvdream,long2023wonder3D}, we extend the original 2D self-attention layer of Stable Diffusion into 3D by concatenating the keys and values of different views within the self-attention layers, which facilitates the interactions across different input views and implicitly encodes multi-view correspondence. Additionally, camera poses are injected into the denoisinge U-Net for both the conditional multi-view inputs and the target view. Specifically, a two-layer MLP is employed to encode camera poses into one-dimensional embeddings, which are then added to time embeddings as residuals.

\subsection{Data Augmentation}
\label{sec:RT}
We train the multi-view conditioned diffusion model with ground-truth multi-view images rendered from Objaverse~\cite{deitke2023objaverse} as condition. However, directly training with these ground-truth images can lead to suboptimal results during inference, as the domain discrepancy between the ground-truth multi-view images and the synthesized ones used during testing  leads to artifacts and inconsistent generation results. To  empower the
model to extract strong 3D priors from inconsistent multi-view inputs at test time, we introduce several data augmentation strategies during the training stage: 

\begin{itemize}
    \item \textit{Noise Disturb and Random Scale}: To imitate the blurry textures and local inconsistencies present in synthesized  multi-view images, we introduce noise disturb augmentation. Specifically, we perturb the clean conditional multi-view images with random Gaussian noises,  which is similar as the forward diffuse process. During our experiments, we adopt the DDPM forward strategy and uniformly sample noises under $t=300$ to destroy the conditional mutli-view images, which are then used as conditional inputs to train the multi-view conditioned diffusion model. To further enhance the networks' robustness towards blurry conditional inputs, we propose a random scale augmentation strategy, which employs a downsample-and-upsample approach to generate blurry training inputs. 
    \item \textit{Random Drop}: During training, we observed that the model tends to learn strong bias by relying heavily on the views closest to the target view for generating novel views, while ignoring the influence of other conditional views. This leads to artifacts and instable results as errors may exist in the closest conditional views. To eliminate this tendency, we introduce random drop augmentation, by randomly  dropping different conditional views during training, enforcing the network to synthesize more consistent results by integrating information from all available views.
\end{itemize}

Similar as LGM~\cite{tang2024lgm}, we also incorporate grid distortion to disrupt the 3D coherence of ground-truth images and camera jitter to vary the conditional camera poses of multi-view inputs.

To simulate the test scenario, we set the first sampled view as the default input image (simulating user inputs at test stage) and only apply data augmentation to the other conditional views. To further control the influence of different views, we introduce a control label enabling manual adjustment of the  conditioning strength for different conditional views. This label, indicative of the conditional strength, is processed by a two-layer MLP into a one-dimensional vector, which is then combined with the time embedding in the local feature extractor to guide the generation process. 
During training stage, When employing larger augmentation scales, such as noise perturbation with larger timesteps, we assign a lower value to the condition label to reduce the control weights, and vice versa.



\begin{figure*}[tb]
  \centering
    \includegraphics[width=0.95\linewidth]{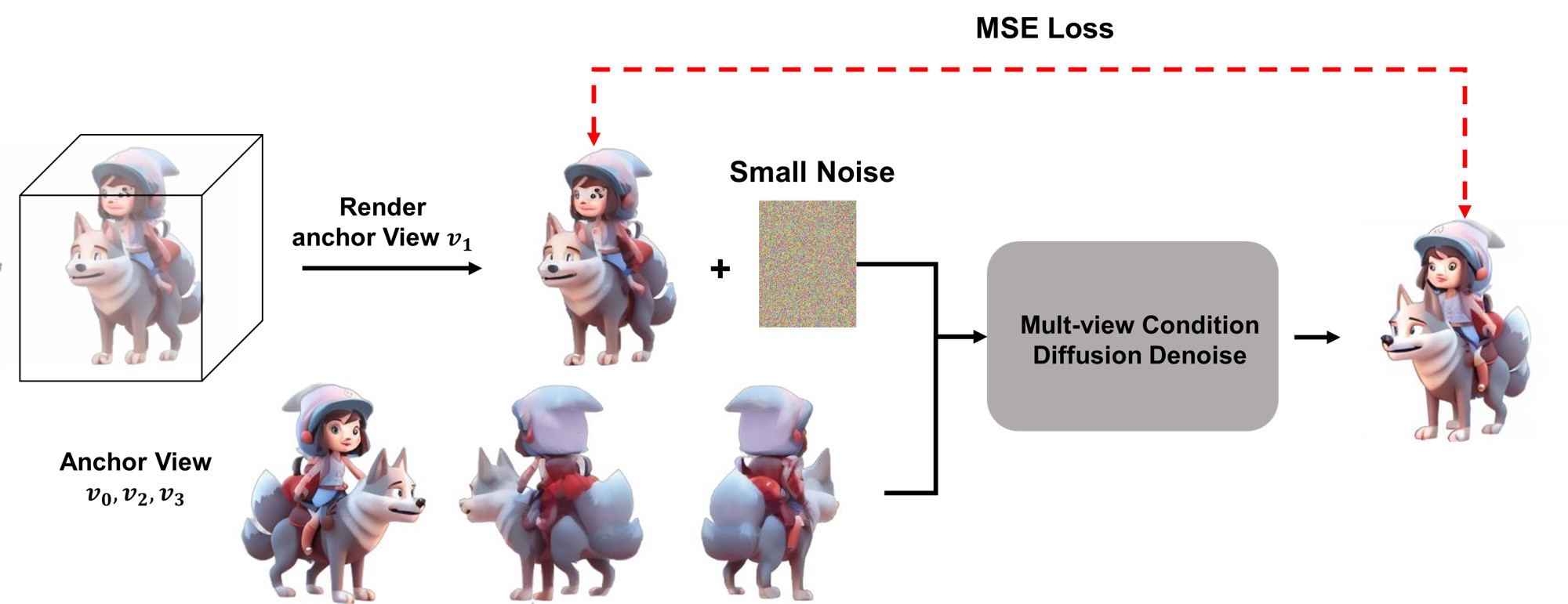}
  \caption{Illustration of the anchor iterative update loss. In detail, we regard the input pesudo multi-view inputs as our initial anchor datasets and adopt an update strategy by first rendering anchor view image, perturbing the image with random noise and then apply a multi-step denoising process with the proposed multi-view condition diffusion model to refine the anchor images. The refined anchor images are then used to supervise the generation with MSE loss to eliminate the over-stature problem during the SDS optimization process~\cite{poole2022Dreamfusion}.}
      \label{fig:du}
\end{figure*}

\subsection{Refinement with SDS Optimization}
\label{sec:Refine}
As shown in Figure~\ref{fig:G_pipeline}, We build our test pipeline with Instant3D~\cite{li2023instant3d}, 
a two-stage feed-forward generation method which firstly generates four multi-view images with finetuned 2D diffusion networks and then lift the multi-view images to 3D model utilizing large reconstruction model. However, due to limited resolution and local inconsistency, the generated results are still in low quality without detailed texture and complicated geometry. 
Leveraging the generated pseudo multi-view as input, we propose to adopt SDS optimization with small noise level to further enhance the coarse generation results. 
Benefit from the high-fidelity geometry and texture information provided by the pesudo multi-view inputs, our model is capable of generating highly consistent novel views and providing precise SDS guidance to enhance the coarse generated results within a short time period ($\sim15$min).

Specifically, we first convert the generated mesh from Instant3D into differentiable 3D representations by randomly rendering and distilling the appearance and occupancy of the mesh with L1 loss. In our experiments, This process is achieved with a fast NeRF (e.g. InstantNGP~\cite{muller2022instant}) with little time cost ($\sim1$min). After initialization, we optimize the fast NeRF utilizing SDS loss with a small range of denoising timestep as [0.02, 0.5]. The optimization process takes about 15min with 2500 steps, which is much more efficient compared to the $1\sim2$ hours time cost of traditional SDS-based methods~\cite{poole2022Dreamfusion, wang2024prolificdreamer, wang2023imagedream}.

We further introduce an Anchor Iterative Update loss to alleviate the over-saturation problem of SDS optimization. Specifically, we draw inspiration from the recent image editing methods~\cite{haque2023instruct,meng2021sdedit} that edit the NeRF by alternatively rendering dataset image, updating the dataset and then supervising the NeRF reconstruction with the updated dataset images. In our generation task, we could also regard the input pesudo multi-view inputs as our initial anchor datasets and adopt a similar update strategy by first rendering anchor view image, perturbing the image with random noise and then apply a multi-step denoising process with the proposed multi-view condition diffusion model to refine the anchor images.
As the denoising process is guided by the multi-view inputs themselves, simply adopt such as process leading to minor refinement on the input anchor images. To address this problem, while updating certain anchor view $v_1$, we drop it in the conditional inputs, and denoise leveraging other views $v_0, v_2, v_3$. We found this works well and achieve accurate refinement on the local details of the anchor image, while preserving the 3D consistency. The refined anchor image is then used to supervise the generation with MSE loss, as illustrated in Figure~\ref{fig:du}. As shown in the bottom line of Figure~\ref{fig:abla}, the proposed Anchor Iterative Update loss alleviates the over-saturation problem and generates realistic textures, leading to better generation performance.

\section{Experiments.}

\subsection{Implementation Details}
\label{sec:imple}



\noindent\textbf{Training.} 
We train our model on 32 NVIDIA V100 GPUs for 30k steps with batchsize 512. The total training process cost about 6 days. We use  $256 \times 256$ as the image resolution for training. For each batch we randomly sample 4 views as conditional multi-view images and another one view as the target view. The model is initialized from MVDream (the version of stable diffusion 2.1) and the optimizer settings and $\epsilon$-prediction strategy are retained from the previous setting for finetuning except we reduce the learning rate to 1e-5 and use 10 times learning rate for camera encoders' parameters for faster convergence.  We train our model on the public available Objaverse~\cite{deitke2023objaverse} dataset following the setting of MVDream~\cite{shi2023mvdream}. Please refer to Appendix for more implementary details.

\noindent\textbf{Testing.} 
We adopt Instant3D~\cite{li2023instant3d}, one of the SOTA 3D generation methods that involves multi-view generation and fast reconstruction as our baseline model to provide the multi-view priors and coarse reconstruction results.  As there is no official code for instant3D~\cite{li2023instant3d}, we reproduce it follows the published paper. However, as discussed in Section~\ref{sec:ablation}, our model could also be plugged into other models capable of providing pseudo multi-view priors, such as InstantMesh~\cite{xu2024instantmesh}, LGM~\cite{tang2024lgm} and etc.



\begin{figure*}[tb]
  \centering
    \includegraphics[width=\linewidth]{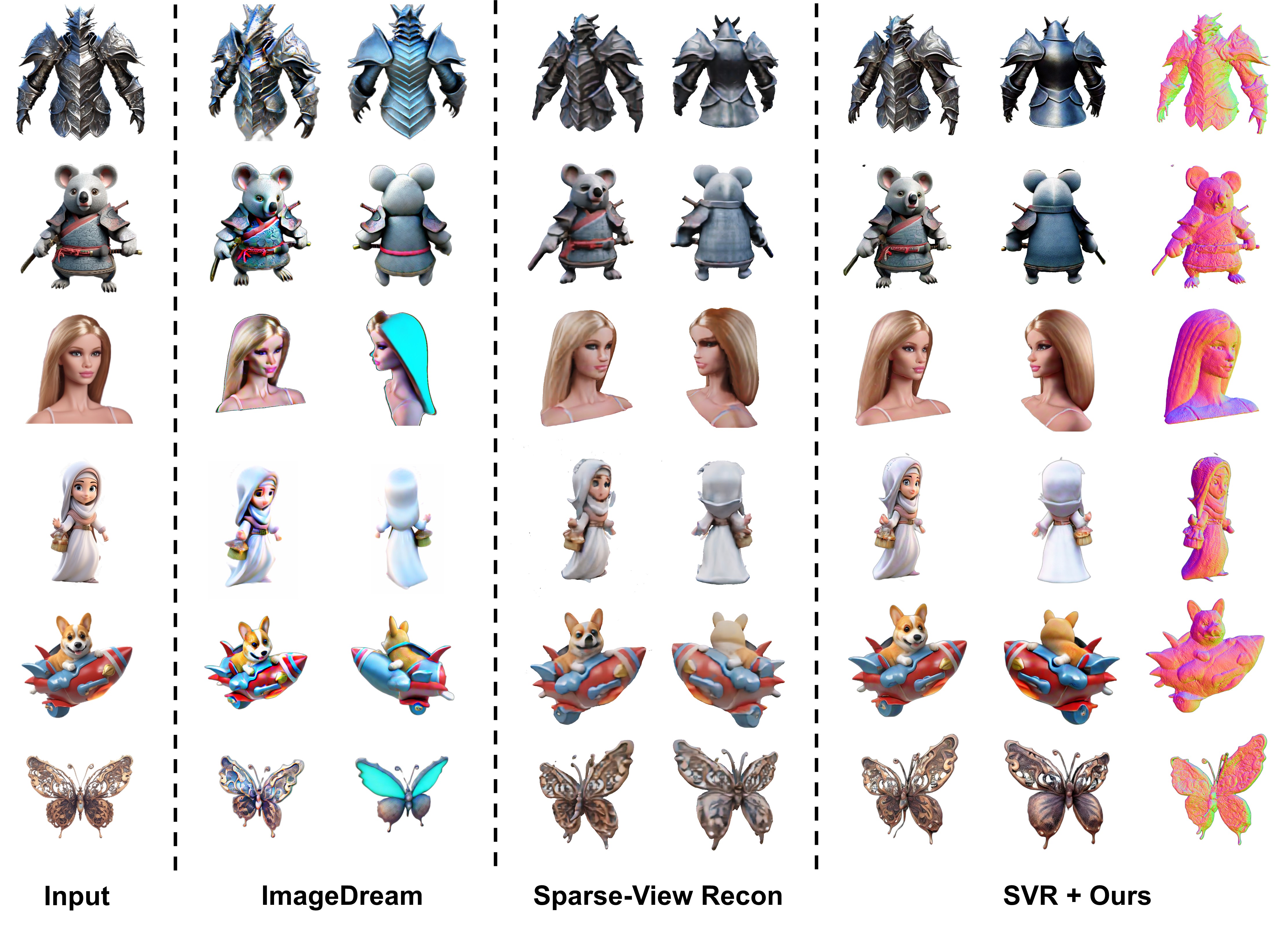}
  \caption{Qualitative Comparison between our method with Imagedream~\cite{wang2023imagedream} and base sparse-view reconstruction model~\cite{li2023instant3d}. SVR denotes Sparse-View Reconstruction.}
\label{fig:Qualtitative}
\end{figure*}

\begin{figure*}[tb]
  \centering
    \includegraphics[width=\linewidth]{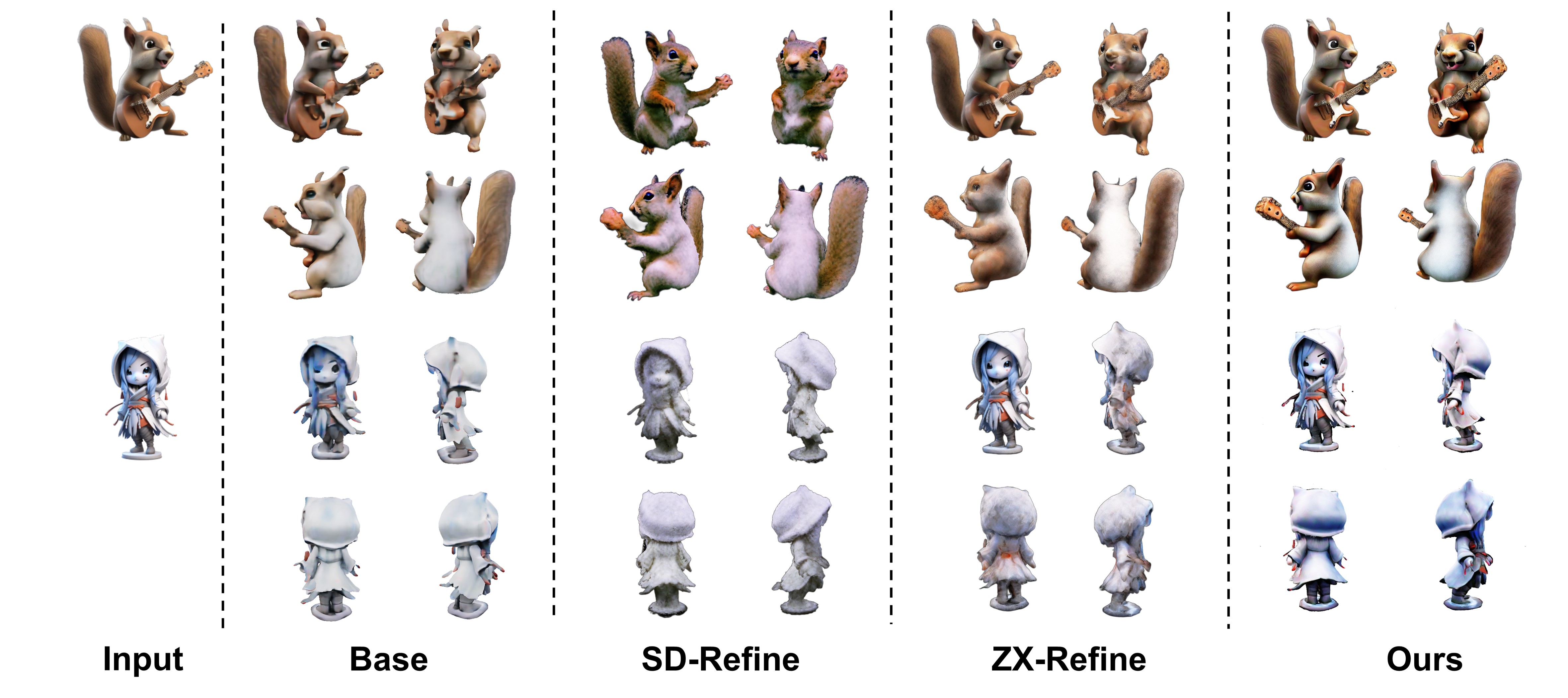}
  \caption{Qualitative comparison between our method with other refinement methods. From left to right, we show input, followed by the reconstruction results from sparse view reconstruction models~\cite{li2023instant3d}, and the results refined by StableDiffusion~\cite{rombach2022high}, Zero123-XL~\cite{liu2023zero}, and Ours.}
\label{fig:boostc0}
\end{figure*}

\subsection{Evaluation on Image-to-3D Generation}
\subsubsection{Qualitative Comparisons.}
We make comparisons with State-Of-the-ArT methods on Image-to-3D generation task, ImageDream~\cite{wang2023imagedream}. As shown in Figure~\ref{fig:Qualtitative},
our method generates visually better results with sharper textures, better geometry and highly consistent 3D alignment. 
ImageDream~\cite{wang2023imagedream} extends MVDream~\cite{shi2023mvdream} to start from a given input image, which proposes a new variant of image conditioning. 
However, limited by the incomplete information provided by a single-view input, it suffers from high uncertainty on the occluded regions, struggling to generate the unseen regions with implausible shapes and chaotic texture. 
In contrast, our model benefits from much stronger and more precise 3d prior provided by the pseudo multi-view inputs, leading to higher generation quality with photo-realistic colors and more geometric details.


We also make comparisons by using other diffusion models at the optimization stage to refine the sparse view reconstruction results, including Stable Diffusion~\cite{rombach2022high} and Zero123-xl~\cite{liu2023zero}. 
As ImageDream~\cite{wang2023imagedream} is trained in its own camera canonical space, it could not be used in the refinement stage directly. For the fairness, we replace our multi-view conditioned diffusion model with other diffusion models at the refinement stage while keeping other settings unchanged.   As shown in Figure \ref{fig:boostc0} and Table \ref{tab:quan}, limited by the ambiguity of text description, Stable Diffusion fails to keep the identity of the initiated coarse object fixed while changing it into another one with different texture and incorrect geometry. 
Although empowered with single-view image input in the cross-attention to synthesize novel views, Zero123-xl still lacks the ability to generate accurate 3D contents with consistent geometry and detailed texture, especially for the unseen regions. 
Compared to these methods, our model acquires strong ability of generating highly consistent images from the pseudo multi-view images and provides more precise SDS guidance which can effectively maintain the identity and enhance local details in both geometry and texture of initial generation results.

    
\begin{table}[tb]
 \caption{ Quantitative Comparisons with image-to-3D methods.  }
  \centering
          \scalebox{0.8}{
  \begin{tabular}{@{}lllll@{}}
    \toprule
    Model & QIS$\uparrow$ & CLIP(TX)$\uparrow$ & CLIP(IM)$\uparrow$ & Time\\
        \midrule
    Zero123-XL~\cite{liu2023zero} & $27.92\pm 2.41 $ & $ 29.11 \pm 3.65$ & $79.48 \pm 6.73$ & ~$\sim10$min \\
    Magic123~\cite{qian2023magic123} & $25.12 \pm 4.29$ & $ 29.31\pm4.69$ & $82.92 \pm 9.33$ & ~$\sim1$h\\
    ImageDream~\cite{wang2023imagedream} & $25.86 \pm 3.68$  &  $31.53 \pm 3.30 $ & $83.79 \pm 4.25$ & ~$\sim2$h \\
    \midrule
    Baseline model~\cite{li2023instant3d} & $23.16 \pm 2.53 $ & $30.61 \pm 3.22 $ & $ 83.20 \pm 5.45 $ & ~$\sim10$s \\
    SD-Refine & $23.85 \pm 4.37 $ & $32.26 \pm 3.51 $ & $ 78.21 \pm 10.32 $ & ~$\sim15$min \\
    Zero123-Refine~\cite{liu2023zero} & $24.08 \pm 3.31 $ & $31.67 \pm 3.09 $ & $ 84.73 \pm 5.96 $ & ~$\sim15$min\\
    \midrule
    MagicBoost (w/o Aug) & $\mathbf{27.96\pm3.22}$ & $31.43 \pm 2.61 $ & $86.24 \pm 4.66 $ & ~$\sim15$min \\
    MagicBoost (w/o AIU) & $27.69 \pm 3.51$ & $31.50 \pm 2.73 $ & $ 87.21 \pm 4.43 $ & ~$\sim15$min\\
    MagicBoost & $27.81 \pm 3.81$ & $\mathbf{31.57 \pm 2.71} $ & $\mathbf{87.30 \pm 4.45}$ & ~$\sim15$min \\
  \bottomrule
  \end{tabular}
  }

  \label{tab:quan}
\end{table}

\begin{figure*}[tb]
  \centering
    \includegraphics[width=\linewidth]{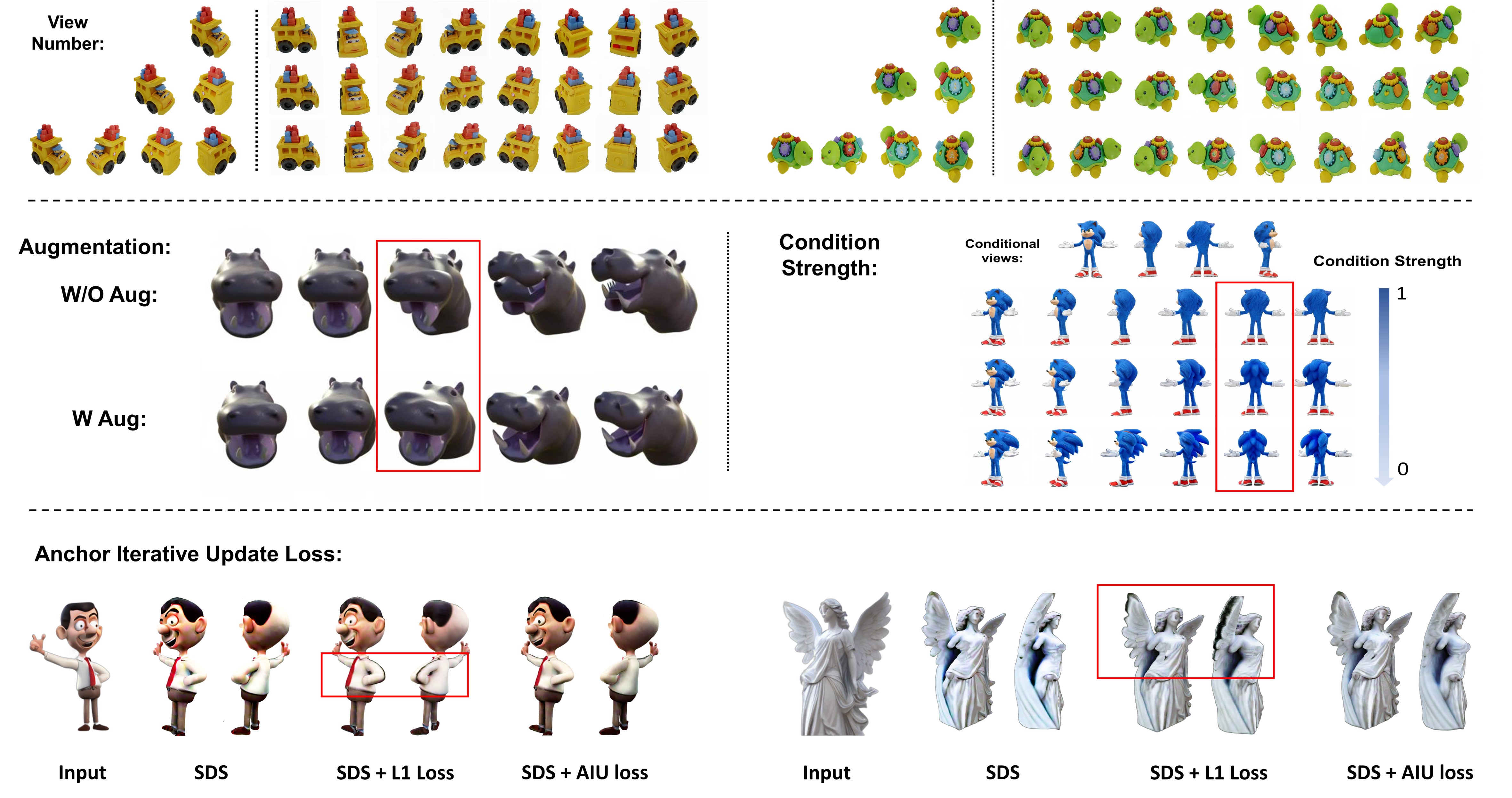}  \caption{Ablation Studies, inluding number of condition views, data augmentation, Anchor Iterative Update loss.}
      \label{fig:abla}
\end{figure*}

\subsubsection{Quantitative Comparisons.}
Following ImageDream~\cite{wang2023imagedream}, we adopt three metrics for quantitative comparison of our methods with others, including the Quality-only Inception Score~\cite{salimans2016improved} (QIS) and CLIP~\cite{radford2021learning} scores that calculated with text prompts and image prompts, respectively. Among which, QIS evaluates image quality and CLIP scores assess the coherence between the generated models and the prompts. The evaluation dataset consists of 37 high-resolution images, which is generated by SDXL with well-curated prompts. We present the quantitative comparison results in Table \ref{tab:quan}. As shown in the results, our model achieves higher scores on all the three metrics compared to other method, demonstrating the great image conditioned generation ability of our models. As the condition image is synthesized by text prompt, the highest CLIP-Text score also demonstrates the capablity of our model to generates highly text-aligned contents. We also compare the inference time of different methods, which is calculated on a single A100 GPU. Notably, the inference speed of our model is significantly faster than the traditional optimization-based methods like Magic123~\cite{qian2023magic123} and ImageDream~\cite{wang2023imagedream}, while achieving better generation quality.

\subsection{Evaluation on Novel View Synthesis}

We employ PSNR, SSIM~\cite{wang2004image}, LPIPS~\cite{zhang2018unreasonable} to evaluate the performance of our model on novel view synthesis task on the Google Scaned Dataset (GSO)~\cite{downs2022google}. In details, we use the same 30 objects chosen by SyncDreamer~\cite{liu2023syncdreamer} and  render 16 views with uniformly distributed camera poses and environment lighting for each object. To ensure a fair and efficient evaluation process, the first render view is selected as the input image for each baseline and our methods. For multi-view inputs of our method, we select views orthogonal to the first input view. As shown in \ref{tab:mv}, benefit from the more comprehensive information provided by multi-view inputs, our model significantly outperforms previous methods that rely on only single-view image as input and synthesize novel views with much higher fidelity. We provide more novel view synthesis results in Figure~\ref{fig:mv}.

\begin{table}[tb]
 \caption{ Quantitative Comparisons in Novel view synthesis. We report PSNR, SSIM~\cite{wang2004image}, LPIPS~\cite{zhang2018unreasonable} on the GSO~\cite{downs2022google} dataset.}
  \centering
  \begin{tabular}{@{}lllll@{}}
    \toprule
    Model & Ref Views & PSNR$\uparrow$ & SSIM$\uparrow$ & LPIPS$\downarrow$ \\
        \midrule
    Realfusion~\cite{melas2023realfusion} &  1 & 15.26 & 0.722 & 0.283 \\
    Zero123~\cite{liu2023zero} & 1 & 18.93 & 0.779 & 0.166 \\
    Syncdreamer~\cite{liu2023syncdreamer} & 1 & 20.05 & 0.798 & 0.146 \\
            \midrule
    Ours & 1 & 20.31 & 0.802 & 0.141 \\
    Ours & 2 & 21.29 & 0.828 & 0.137 \\
    Ours & 4 & 23.98 & 0.862 & 0.105 \\
  \bottomrule
  \end{tabular}
  \label{tab:mv}
\end{table}

\subsection{Ablation Study.}
\label{sec:ablation}


\noindent\textbf{Number of condition views.} 
We show the effects of our multi-view conditioned diffusion model to synthesize novel views with different number of conditional views in Figure~\ref{fig:abla} and Table~\ref{tab:mv}. When there is only one single view input, our model fails to generate consistent novel-view images. This demonstrates that the diffusion model struggle to acquires accurate 3D information from only one view input, resulting in the inconsistent generation results. However, as the number of the condition views increases, the generation fidelity of our model gradually increases, leading to more consistent results which demonstrates the necessity of our proposed multi-view conditioned diffusion model.


\noindent\textbf{Data augmentation.} 
We compare models trained with/without data augmentation in Table \ref{tab:quan} and Figure \ref{fig:abla}. As shown in Figure~\ref{fig:abla}, the Model trained without data augment fails to generate 3D consistent results due to the domain gap between training stage and inference stage. In comparison, with the proposed data augmentation strategy, our model learns to better correct the inconsistency in the conditional views and generate highly consistent 3D results. 
Figure \ref{fig:abla} also illustrates the effect of the proposed control label. By fixing the weights of the first input view while lowing the weights of others, the model learns to generate diverse realistic results without constraining strictly by the input views. By setting the condition strength at different level, our model is capable to deal with inputs in different quality level, which improves the robustness of our model. 


\noindent\textbf{Anchor Iterative Update loss.} 
We adopt Anchor Iterative Update loss to deal with the over-saturation problem of SDS. A more straightforward method is to simply use L1 loss from four pesudo multi-view images to constrain the appearance of the object from being awary from the input. However, as illustrated in the bottom line of \ref{fig:abla}, L1 loss highly rely on the consistency of the input multi-view images, leading to obvious artifacts with collapsed textures at the inconsistent regions between different views. Instead, Anchor Iterative Update losss, leveraging the iterative render, update and distill strategy, alleviates the  over-saturation problem by gradually distilling the clean RGB image refined from the updated rendered anchor images, leading to more robust performance with realistic appearance.


\begin{figure*}
  \centering
    \includegraphics[width=\linewidth]{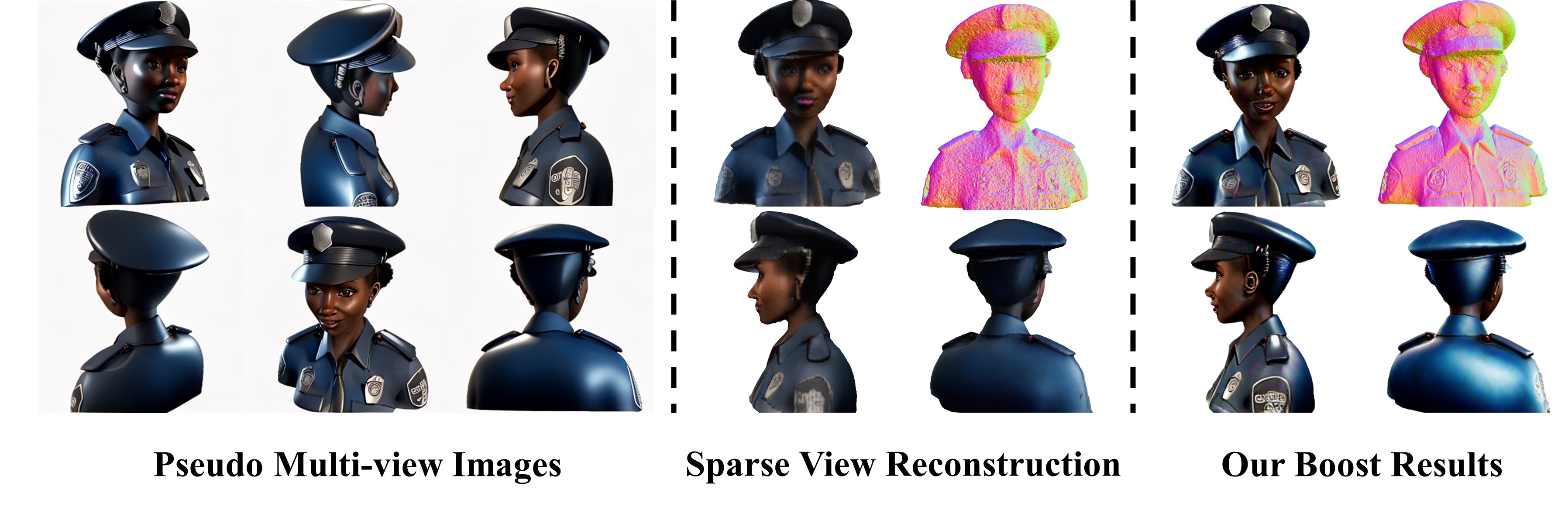}
  \caption{Our model could also be plugged into other models capable of providing
pseudo multi-view priors. In this figure we show boost results of our model pluged into InstantMesh~\cite{xu2024instantmesh}. (See appendix for more visual results).}
\label{fig:imabla}
\end{figure*}

\noindent\textbf{Plug in different 3D generation models.} Although we adopt Instant3D~\cite{li2023instant3d} as our baseline model in our experiments. Our model could also be plugged into other models capable of providing
pseudo multi-view priors, such as InstantMesh~\cite{xu2024instantmesh}, LGM~\cite{tang2024lgm} and etc. Figure~\ref{fig:imabla} shows an example of adopting InstantMesh~\cite{xu2024instantmesh} as our base model and demonstrates the generalization ability of our model for different 3D generation methods. (See appendix for more visual results).

\begin{figure*}
  \centering
    \includegraphics[width=\linewidth]{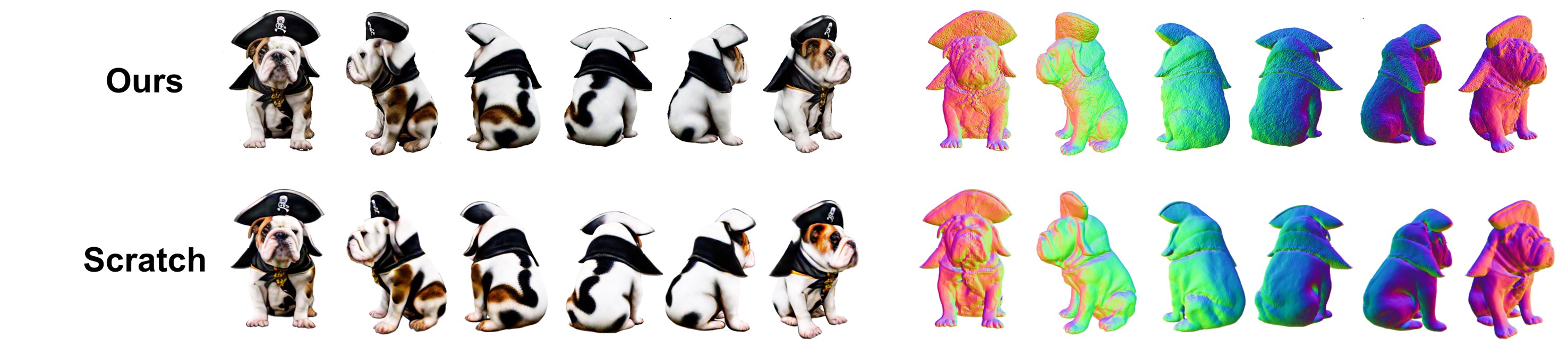}
  \caption{Comparison with generation from scratch. Given the coarse 3D model as initialization, our model is capable to generate high-quality results comparable to those generating from scratch, which takes $\times5$ time ($\sim1.2$h), emphasizing the effectiveness of our overall pipeline.}
\label{fig:scratch}
\end{figure*}

\noindent\textbf{Generation with coarse initialization.} 
In addition, we evaluated the impact of utilizing the coarse reconstruction outcomes obtained from a larger reconstruction model as an initialization step. As illustrates in Figure~\ref{fig:scratch}, when provided with pseudo multi-view inputs, our model is capable of producing high-quality 3D contents from scratch. However, this process requires approximately $\times 5$ time (1.2h) for our model to achieve comparable results to those generated using the coarse reconstruction outputs as initialization. This result emphasizes the effectiveness of our overall approach.



\section{Limitation.}
\label{sec:lim}
Our model can effectively generate high-quality outputs, but its performance is still limited by the following factors: 1) The use of pseudo 3D multi-view images synthesized by the finetined 2D multi-view diffusion model as inputs imposes an unavoidable influence on our model's performance. Therefore, the generation quality of the multi-view diffusion model directly affects our model's performance. Exploring 2D multi-view diffusion models with better generation quality would further improve our model's performance. 2) Our model's resolution $256 \times 256$ remains constrained in contrast to text-conditioned 2D diffusion models, which generate images with higher resolution containing more details. Uplifting our model to a higher resolution, such as $512 \times 512$, would produce superior generation outcomes, allowing for more intricate geometry and detailed texture.

\section{Conclusion}
\label{sec:con}
We present Magic-Boost, a multi-view conditioned diffusion model which takes pseudo generated multi-view images as input, 
capable of synthesising highly consistent novel view images and providing precise SDS guidance during the optimization process.  Extensive experiments demonstrate our model greatly enhances the generation quality of coarse input and generates high-quality 3D assets with detailed geometry and realistic texture within a short time period.

{\appendix[Additional Implementation Details]
\noindent\textbf{Network Structure.}
We build our network on the Stable Diffusion backbone~\cite{rombach2022high}, while we make several changes on the structure: 1) We incorporate a frozen CLIP pre-trained Vision Transformer~\cite{radford2021learning} as global feature extractor. 2) We adopt the same denoising U-Net with shared weights to extract local features at fixed time step $0$.
3) We extend the original 2D self attention layer into 3D by concatenating keys and values from different views to facilitate information propagation. 4) We encode the camera pose with a two layer MLP, which is then added to time embedding as residuals. The proposed control label is embedded and added to the time embedding in the same way. 

\noindent\textbf{SDS Optimization.}
We use the implicit volume implementation in threestudio~\cite{guo2023threestudio} as our 3D representation,
which includes a multi-resolution hash-grid and a MLP to predict density and RGB. For camera views,
we sample the camera in exactly the same way as how we render the 3D dataset. To optimize the coarse generation results, we first convert the generated mesh from Instant3D into the implicit volume by randomly rendering and distilling the appearance and occupancy of the mesh with L1 loss. We set the the total steps for this stage to be 1000, which takes about 1 min.  After this initialization stage, we optimize the implicit volume utilizing SDS loss with a small range of denoising timestep as [0.02, 0.5]. The optimization process takes about 15min with 2500 steps. The update interval for the proposed Anchor Iterative Update loss is set to be 500 and the control label is set to 1 for the first input view and 0.5 for the other three input views in our experiments. Similar as prior arts~\cite{qian2023magic123,melas2023realfusion}, we adopt a l1 loss from the first input view to further refine the details in the generation results.  In both stages, we adopt Adam optimizer and adopt the orientation loss~\cite{poole2022Dreamfusion} to enhance the performance.

\begin{figure*}[tb]
  \centering
    \includegraphics[width=\linewidth]{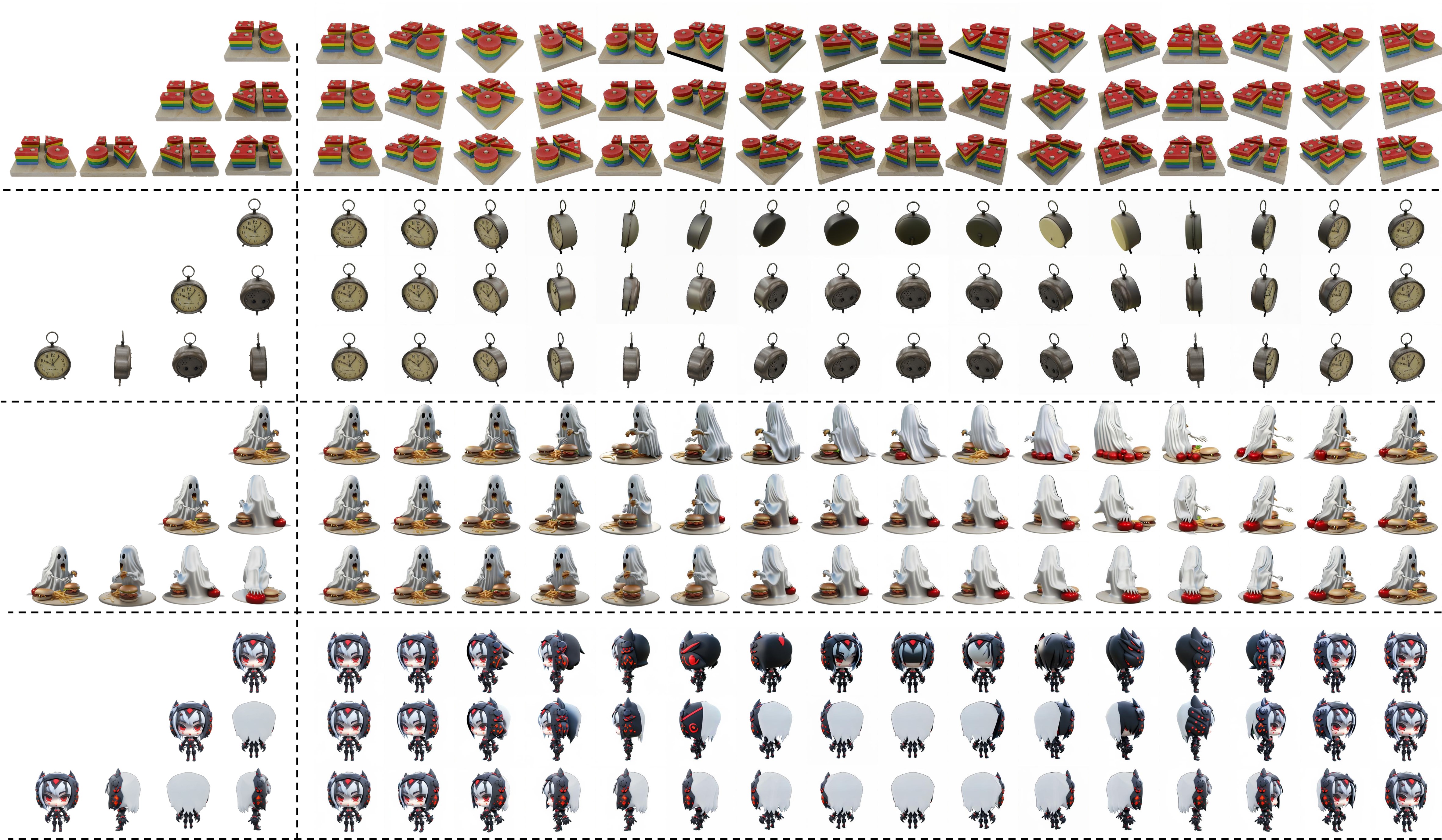}
  \caption{Effects of using different condition views as input. Left column shows the input multi-view images and the right column shows synthesized novel views.}
\label{fig:mv}
\end{figure*}

\noindent\textbf{Instant3D Implementation.}
We reproduce Instant3D~\cite{li2023instant} following the published paper while making several minor changes: 1) We adopt ImageDream~\cite{wang2023imagedream} and MVDream~\cite{shi2023mvdream} as the multi-view diffusion models at the first stage to generate four multi-view images from single-image and text prompt, respectively. 2) We implement the sparse-view large reconstruction model strictly follow the original paper while training it on the same dataset as the proposed Magic-Boost method. The whole training procedure takes about 7 days on 32 A100 GPUs.}

\bibliographystyle{IEEEtran}
\bibliography{reference.bib}

\clearpage

\section{
More Qualitative Results.}
\label{sec:MQ}
\begin{figure*}[t]
  \centering
    \includegraphics[width=0.7\linewidth]{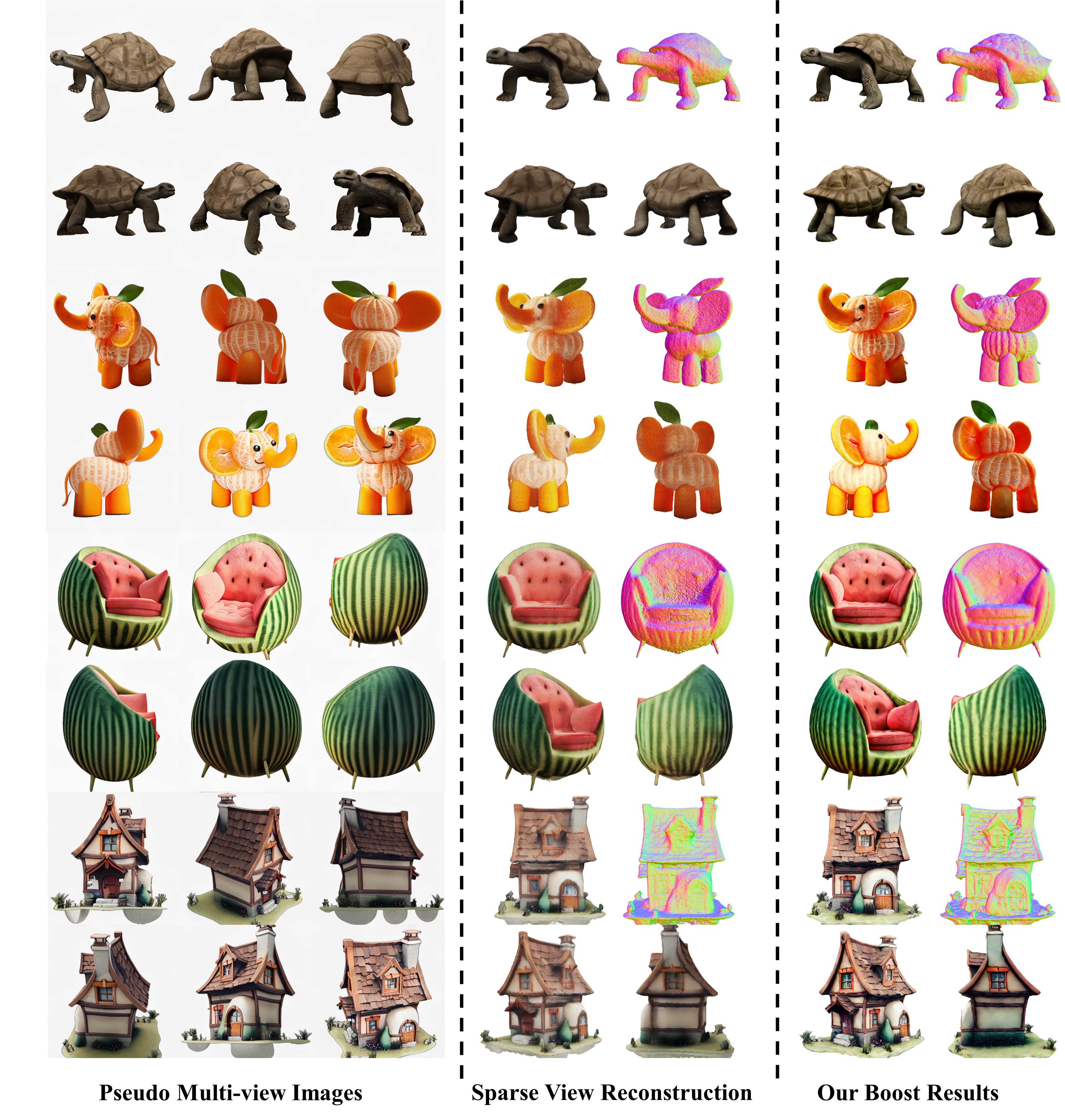}
  \caption{More results of Image conditioned generation based on InstantMesh~\cite{xu2024instantmesh} From left to right: the input image, the pseudo multi-view images, the sparse view reconstruction results and the boosted results of our method.}
\label{fig:mesh}
\end{figure*}

Please refer to our project page: \href{https://magic-research.github.io/magic-boost/}{project page} for video results and more qualitative comparisons and our GitHub for code: \href{https://github.com/magic-research/magic-boost}{code}. 
Figure~\ref{fig:mesh} shows more results our method based on InstantmMesh~\cite{xu2024instantmesh}. 

\vfill

\end{document}